\documentclass{article} 
\usepackage{iclr2019_conference,times}

\usepackage{amsmath,amsfonts,bm}

\def\eqref#1{equation~\ref{#1}}

\def\1{\bm{1}}

\DeclareMathAlphabet{\mathsfit}{\encodingdefault}{\sfdefault}{m}{sl}
\SetMathAlphabet{\mathsfit}{bold}{\encodingdefault}{\sfdefault}{bx}{n}

\usepackage{hyperref}
\usepackage{url}
\usepackage{times}
\usepackage{soul}
\usepackage{url}
\usepackage[utf8]{inputenc}
\usepackage[small]{caption}
\usepackage{graphicx}
\usepackage{amsmath}
\usepackage{booktabs}
\usepackage[algo2e]{algorithm2e}
\usepackage{algorithm}
\usepackage[noend]{algpseudocode}
\usepackage{lipsum}
\usepackage{mwe}
\title{A Method for Computing Class-wise \\Universal Adversarial Perturbations}

\author{Tejus Gupta \thanks{authors contributed equally},  Abhishek Sinha$^{*}$\\
Work done at Adobe \\
\texttt{\{tejusgupta14,abhishek.sinha94\}@gmail.com} \\
\AND
Nupur Kumari$^{*}$, Mayank Singh$^{*}$ \& Balaji Krishnamurthy\\
Adobe Inc.,Noida\\
\texttt{\{nupkumar,msingh,kbalaji\}@adobe.com} \\
}

\iclrfinalcopy 
\begin{document}

\maketitle

\begin{abstract}
  We present an algorithm for computing class-specific universal adversarial perturbations for deep neural networks. Such perturbations can induce mis-classification in a large fraction of images of a specific class. Unlike previous methods that use iterative optimization for computing a universal perturbation, the proposed method employs a perturbation that is a linear function of weights of the neural network and hence can be computed much faster. The method does not require any training data and has no hyper-parameters. The attack obtains 34\% to 51\% fooling rate on state-of-the-art deep neural networks on ImageNet and transfers across models. We also study the characteristics of the decision boundaries learned by standard and adversarially trained models to understand the universal adversarial perturbations.
\end{abstract}

\section{Introduction}
The vulnerability of state-of-the-art neural networks to adversarial perturbations was first studied in \citep{szegedy2013intriguing}. Adversarial perturbations are visually imperceptible modifications to the input of neural networks that results in abrupt change in output. Subsequent works \citep{goodfellow2014explaining, moosavi2016deepfool, carlini2017towards, madry2017towards} have introduced stronger attacks that induce more mis-classifications with perturbations of smaller norm. Adversarial examples suggest that while CNNs perform remarkably well, they may not be learning interpretable and robust features (as discussed by \citep{tsipras2018there}).

Another line of research focuses on adversarial attacks that pose a serious security threat in real-world systems. Adversarial examples transfer across different models and this property enables practical black-box attacks \citep{papernot2017practical, liu2016delving, alp2018broken}. Machine learning systems that operate in the physical world with inputs from different sensors have also been shown to be vulnerable to such attacks \citep{kurakin2016adversarial}. \citep{bhagoji2017exploring,ilyas2018prior, ilyas2018black} have also shown that adversaries with a limited number of query access to model and information about the prediction of top labels can also construct adversarial examples.

Most of the proposed defenses against adversarial examples have been subsequently broken or had its efficacy severely reduced   \citep{uesato2018adversarial, athalye2018obfuscated}. Currently, adversarial training with PGD \citep{madry2017towards} perturbation is the state-of-the-art defense against adversarial examples . However, this defense is computationally expensive and isn’t robust to sets of perturbations with different norm constraint that weren’t used during adversarial training. Recent work has focused on certified defenses that computes a provable upper bound on the adversarial perturbation against which the model is robust and try to maximize it \citep{kolter2017provable, raghunathan2018certified, wong2018scaling, provable_certificate4}. 

We present a novel method for constructing class-wise universal adversarial perturbations that can induce mis-classification for a large fraction of images of a particular class. The attack is particularly interesting due to its simplicity. It can be crafted using a linear function of the trained weights of deep neural networks. 
The technique doesn't require access to the training dataset and adversarial perturbations for the deep learning model can be calculated within seconds. To the best of our knowledge, our approach is the first work on computing (class-wise) universal adversarial perturbations on deep neural networks without any iterative optimization.

The main contributions of this paper are as follows:
\begin{enumerate}
\item We propose a data-independent method for generating class-wise universal adversarial perturbations. The proposed perturbation exploits the linearity of the decision boundaries of deep neural networks. Since the proposed algorithm isn't optimization-based and has a closed form formula, it can generate universal perturbation much faster than previous methods \citep{mopuri2017fast,Moosavi-Dezfooli_2017_CVPR}.
\item We evaluate the performance of the proposed method on models trained over MNIST \citep{mnist_dataset}, CIFAR-10  \citep{cifar_dataset} and ImageNet  \citep{russakovsky2015imagenet} datasets. We also find that the attack has high cross-model generalizability.
\item We find that the directions of example-specific adversarial perturbations are aligned with our universal perturbation. We conduct experiments to support the hypothesis that class boundaries in neural networks are largely linear. We also find that some classes in ImageNet dataset are inherently harder to misclassify using universal adversarial perturbations.
\end{enumerate}



\begin{figure*}[t]
\begin{center}
\includegraphics[width=1.25in,height=1.25in]{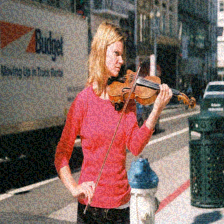}
\hspace{0.1cm}
\includegraphics[width=1.25in,height=1.25in]{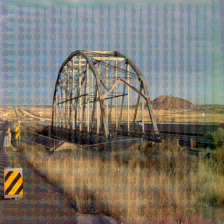}
\hspace{0.1cm}
\includegraphics[width=1.25in,height=1.25in]{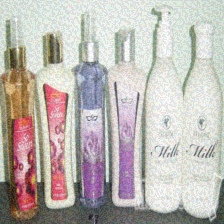}
\hspace{0.1cm}
\includegraphics[width=1.25in,height=1.25in]{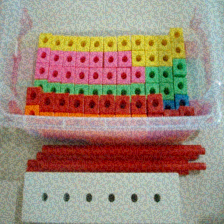}
\end{center}

\begin{center}
    \caption{Examples of images after adding adversarial perturbation. Label of clean images (left to right): violin, steel arch bridge, lotion, abacus. Predicted Label of adversarial images (left to right): flute, racket, carpenter's kit, mouth-harp}
\end{center}
\label{fig:exampleperturb}
\end{figure*}

\section{Related Work}
Recently, \citep{Moosavi-Dezfooli_2017_CVPR} demonstrated that state-of-the-art neural networks are vulnerable to image-agnostic adversarial perturbations. They start with a perturbation $v = 0$ , and proceed iteratively over a subset of the training dataset. At each iteration, they compute the smallest $\Delta v$ such that $v+\Delta v$ induces mis-classification in the current batch of images and modify $v = v+\Delta v$. To ensure that $||v||_p < \epsilon$ is satisfied, they project $v$ on the $\ell_p (\text{where } p \in [1,\infty])$ ball of radius $\epsilon$ with center $0$ at each iteration. They generate adversarial perturbation through this algorithm and found that it generalized to the validation dataset. Since the perturbations also generalize well across deep neural networks, an adversary can generate universal perturbations for a model even in a black-box setting.

In \cite{moosavi2017analysis}, Moosavi-Dezfooli et~al. derive the sufficient conditions on decision boundaries for the existence of universal adversarial perturbations, and these conditions are empirically verified for deep networks. The analysis shows that it is necessary to suppress the space of shared positive adversarial directions for robustness against universal adversarial perturbations. 


\citep{art_singular_2018_CVPR} present a method for constructing universal adversarial perturbations on ImageNet models using only 64 images. There has also been work on generating universal adversarial perturbations through a generative model \citep{poursaeed2018generative, hayesuniversal, reddy2018nag}.
For a comprehensive review of the work done in the area of adversarial examples, please refer \citep{reviewpaper,reviewpaper2}.

The paper closest to our work is \citep{mopuri2017fast}, where the authors propose a method of generating universal perturbations that is independent of training data. They start with a random perturbation and iteratively optimize to over-saturate the features at intermediate convolution layers in the target model. Their optimization objective is:

\begin{equation} \label{eq1}
L = -log(\prod_{i=1}^{K} l_i( \Delta \vec{x}) ) \textnormal{ such that } ||\Delta \vec{x}||_{\infty} < \epsilon
\end{equation}

where $l_i(\Delta \vec{x})$ is the average activation of $i^{th}$ layer on input $\Delta \vec{x}$, $K$ is the number of CNN layers over which they do the optimization and  $\epsilon$ is the max-norm constraint.

\vspace{0.5cm}

However,  our  paper is different from \citep{mopuri2017fast} on the following counts:

\begin{enumerate}
\item In \citep{mopuri2017fast}, the perturbation is not optimized for eq.\ref{eq1} till convergence. Instead, they use a held-out set and terminate the optimization process when the fooling rate on this set stops increasing. Our method doesn't use any validation data for computing adversarial perturbation.
\item Unlike \citep{mopuri2017fast}, we don't use iterative optimization because of which our method is much faster. Additionally, the absence of hyper-parameters in our method precludes any fine-tuning of hyper-parameters.
\item A limitation of our work is that the adversary needs to know the ground truth class of clean examples. The validity of this assumption depends on the application, but we would like to point out that the adversary can use our perturbation for stopping the model from predicting a particular class in any scenario. For example, the adversary can stop the network from predicting class $C_i$ by always adding the perturbation for class $C_i$ regardless of the class of clean image.
\end{enumerate}\section{Methodology}
In this section, we formalize the notion of class wise universal perturbations and propose a method for crafting such perturbations. Let $\psi{_i}$ denote a distribution of images of class $i$ in $\mathbb{R}^{d}$, and $C$ be a neural network classifier with $k$ classes that outputs for each image $\vec{x} \in \mathbb{R}^{d}$ an estimated class $C(\vec{x})$ (where $C(\vec{x}) \in \{1,2, ...,k\}$). We aim to construct perturbation vectors $\Delta \vec{x_i} \in \mathbb{R}^{d}$ for each class $i$ (where $i \in \{1,2, ...,k\}$) that fools the classifier $C$ on almost all data-points sampled from $\psi{_i}$. That is we require $\Delta \vec{x_i}$ such that
\begin{equation}\label{eq4}
    \begin{aligned}
    &C(\vec{x} + \Delta \vec{x_i} ) \neq C(\vec{x}) \; \text{for } \;\vec{x} \sim \psi_{i} \\
    &and\;||\Delta \vec{x_i}||_{\infty} \leq \epsilon 
    \end{aligned}
\end{equation}

Our construction of adversarial perturbations for deep neural networks is motivated from the optimal adversarial attack in linear classifiers. Consider a linear/affine classifier $F$ with $k$ classes, such that:
\begin{equation}\label{eq5}
    \begin{aligned}
    &F(\vec{x}) = \underset{i}{\arg\max} \hspace{0.2cm} f_i(\vec{x})\\
    &f_i(\vec{x}) =  \vec{w_i^T}\vec{x} + \vec{b_i}\; | \; i \in \{1, 2, ... , k\}
    \end{aligned}
\end{equation}
  For any input $\vec{x}$ with $F(\vec{x}) = l$, $F$ outputs the same label inside a convex polyhedron $S = \{\vec{x} | f_l(\vec{x}) >= f_i(\vec{x}), i \in \{1, 2, ... , k\}\}$. The distance to the decision boundary separating class $l$ and class $i$ is
  \begin{equation} \label{eq6}
\begin{split}
dist_i(x) = \frac{\vert f_i(\vec{x}) - f_l(\vec{x}) \vert}{  || \vec{w_i} - \vec{w_l}||_1}
\end{split}
\end{equation}
  
  The optimal direction of adversarial perturbation \citep{moosavi2016deepfool} is towards the closest decision plane of $F$. Therefore the optimal adversarial perturbation within max norm constraint($||\Delta \vec{x}||_{\infty} < \epsilon$) is,   

\begin{equation} \label{eq7}
\begin{split}
\Delta \vec{x} &= \epsilon(sign(\vec{w_{t(x)}} - \vec{w_{l}})) \\
\textnormal{where } t(\vec{x}) &= \underset{i \neq l}{\arg\min}\;dist_i(\vec{x})
\end{split}
\end{equation}

In our work, we consider a semi-universal attack where the same adversarial perturbation can induce mis-classification for a large fraction of examples in a particular class. To construct our data-independent perturbations in the case of deep neural networks we incorporate two approximations over the above discussed linear perturbation.

\begin{enumerate}\label{approx}

\item [A1. \label{approx_a1}]
The perturbation $\Delta \vec{x}$ in eqn. \ref{eq7} is the optimal perturbation for minimizing the difference between logit value of predicted class and target class i.e. $f_{l}(\vec{x})-f_{t(x)}(\vec{x})$. As our attack is independent of input $x$, we can't find the best target class $t(x)$. So we approximate by minimizing the difference between the logit value of predicted class and average of other classes. This is equivalent to replacing $\vec{w_{t(x)}}$ by $\sum_{\substack{i \in \{1,2,..k\}, i \neq F(x)}} \vec{w_i} / (k-1)$ in eqn. \ref{eq7}.\\

\item [A2.]\label{approx_a2} Neural networks are a composition of linear layers and non-linear activations. For a given neural network, we create a second network with the same weights and architecture but ignoring the activation function, and replacing max-pool layers by average-pool layers. The adversarial perturbation is then computed for this linear network and is used to attack the original network. 

Given a $n$ layered neural network classifier $F$ with parameters ${W_j}$, $b_j$ and non-linear activation functions $a_j$ ($j \in \{1,2,..n-1\}$). For a data point $x$, $F(x)$ and its corresponding linear network i.e. $F_{linear}(x)$ are as follows:
\begin{equation}\label{eq8}
    \begin{aligned}
    F(x) &= W_n \times a_{n-1}(W_{n-1} \times  (.... (W_2 \times a_1(W_1\vec{x} + b_1)\\ & \qquad + b_2) ....) + b_{n-1}) + b_n
    \end{aligned}
\end{equation}
\begin{equation}\label{eq9}
    \begin{aligned}
    F_{linear}(x) &= W_n \times (W_{n-1} \times (.... (W_2 \times (W_1\vec{x} + b_1)\\ & \qquad  + b_2) ....) + b_{n-1}) + b_n \\
    &= U\vec{x} + b 
    \end{aligned}
\end{equation}
\end{enumerate}
We obtain class-wise universal adversarial perturbation using the matrix $U$. The universal perturbation for class $j$ can be defined as,

\begin{equation} \label{eq10}
\begin{split}
\Delta \vec{x_j} = \epsilon(sign(\sum_{i \in \{1,2,..K\}, i \neq j} \vec{u_i} / (K-1) - \vec{u_{j}})) \\
\end{split}
\end{equation}

Here $\vec{u_i}$ denotes the $i_{th}$ row of matrix $U$. We demonstrate the effect of both the approximations on the performance of attack in Section 3.

\textbf{Implementation details\footnote{For reproducibility of our results, we will open-source our code after publication.}:} We can compute the matrix $U$ by multiplying weight matrices of each linear layer. But this method is computationally expensive as the weight matrices corresponding to convolutional layers are large. Instead we use eq. 9 to compute $U$ efficiently. For a randomly sampled $\vec{x}$ from $\mathbb{R}^{d}$, $U$ is given by : 
\begin{equation}
 U = \frac{\partial F_{linear}(x)}{\partial x}
\end{equation}\label{eq9}

\textbf{Threat model:} If an adversary can generate universal adversarial perturbations, it can be used to attack a high volume of data at a low computational cost. Class-specific adversarial perturbations may be used to prevent the system from classifying the input as a particular class. For example, in the case of classifying inappropriate/nude/NSFW images, the adversary can modify the image to get it classified as non-NSFW.

Our attack allows the adversary to compute adversarial perturbations without having access to training data. Even if the adversary has access to training data, it may choose not to use the training data to improve cross-dataset transferability \citep{mopuri2017fast}. 

\begin{algorithm}[H]
\SetAlgoLined
\Begin{
\footnotesize{
    \textbf{Input}: Neural Net Classifier $F$ with $k$ classes.
    \textbf{Output}: Matrix $U$. \\ 
    Construct $F_{linear}$ from $F$\\
    $d = $ Input dimension of image \\
    Randomly sample $x$ from $\mathbb{R}^{d}$ \\
    $y = F_{linear}(x)$ \\
    \For{$i \in {1, 2, ..., k}$}
    {
        $ y[i].backward();$\\
        $U[i,:] =$ x.grad\\
    }
    \textbf{return} $U$. 
 }
 }
 \caption{\footnotesize{Pseudo-code for Computing Universal Matrix}}
\end{algorithm}

\begin{table*}
\centering
\begin{tabular}{lrrrrr}  
\toprule
&VGG-F & CaffeNet & Resnet-18 & ResNet-50 & VGG-16 \\
\midrule
VGG-F &\textbf{51.74} & 36.47 & 22.35 & 15.63 & 25.17\\
 CaffeNet &40.63 & \textbf{47.32} & 36.79 & 25.84 & 39.43\\
 Resnet-18 &49.40 & 41.26 & \textbf{43.57} & 33.34 & 46.85\\
 ResNet-50 &46.31 & 40.78 & 41.08 & \textbf{33.82} & 48.10\\
 VGG-16 &44.72 & 42.68 & 36.08 & 23.63 & \textbf{46.00}\\
\bottomrule
\end{tabular}
\caption{\footnotesize{Fooling rate of proposed attack for various models trained on ImageNet. Row corresponds to the model used for calculating perturbation and column corresponds to the attacked model.}}
\label{tab:foolingrates}
\end{table*}



\section{Experiments}

\textbf{Setup:} Here we briefly describe the network architecture used for each dataset.
\begin{itemize}
\item \textbf{MNIST}: we use a neural network with 2 convolutional layers (32C, 3x3 filter, and 64C, 3x3 filter) and 2 fully connected layers (1024 and 10 units) for our experiments on MNIST dataset. We used ReLU activations. We trained the classifier for 15 epochs using SGD and reached a test accuracy of 99.3\%. 

\item \textbf{CIFAR-10}: we trained a 6 layer convolutional neural network (32C 3x3 conv - 64C 3x3 conv - avg-pool - 128C 3x3 conv - avg-pool - 128C 3x3 conv - dropout - 1500 fc - dropout - 10 fc) using SGD for 60 epochs with L2 regularization and obtained 83.6\% test accuracy.

\item \textbf{ImageNet}: we use pre-trained models for VGG-16 \citep{simonyan2014very},
VGG-F \citep{chatfield2014return},
Caffenet \citep{jia2014caffe},
Resnet-18 and Resnet-50 \citep{he2016deep} from Pytorch \citep{paszke2017automatic} and Tensorflow's \citep{abadi2016tensorflow} model zoo.
\end{itemize}

\subsection{Results}

Table \ref{tab:foolingrates} shows the fooling rates of our attack on a variety of network architectures trained on ImageNet. Each row shows the fooling rates for adversarial perturbations computed using that model and used to attack other models, i.e., the non-diagonal entries are fooling rates for black-box attacks. The attack has a fooling rate of 34-51\% on these deep networks and transfers well to other models. E.g., adversarial perturbations computed for Resnet-18 has fooling rates varying from 33.34\% to 49.40\% on various other architectures. For MNIST and CIFAR-10 we obtained the fooling rate of 87.49\% and 51.75\%. Note that the fooling rate for a random perturbation is 8-10\%. Some
of the adversarial examples are shown in Fig. 1. 

\paragraph{Effect of Approximation A1 and A2:}
We analyze the effect of approximations A1 and A2 as explained in section \ref{approx_a1} on a baseline input-specific attack. The baseline attack is based on equation \ref{eq7} by linearizing the neural network around the current input. We use $\epsilon = 0.3$ for MNIST, $\epsilon = 16.0/255.0$ for CIFAR-10, and $\epsilon = 10.0/255.0$ for ImageNet. As shown in Table 2, we observe high fooling rates despite both the approximations.
\begin{figure}[t]
    \begin{minipage}[c]{.5\textwidth}
            \begin{tabular}{|p{1.6cm}||p{2.7cm}||p{1.4cm}||}
            \toprule
            Dataset  & Attack & Success Rate \\
            \midrule
            MNIST        & Baseline  & 97.47      \\
                        & Baseline + A1  & 81.45       \\
                        & Baseline + A1, A2  & 87.49      \\
            \midrule
            CIFAR-10       & Baseline  & 97.45      \\
                        & Baseline + A1  & 83.89       \\
                        & Baseline + A1, A2  & 51.75      \\
            \midrule
            Imagenet (VGG16)     & Baseline  & 99.43      \\
                        & Baseline + A1  & 96.23       \\
                        & Baseline + A1, A2  & 46.00      \\
            \bottomrule
            \end{tabular}
            \begin{center}
            \caption{\footnotesize{Effect of approximations A1 and A2 on attack success rate}}
            \end{center}
            \label{tab:booktabs}
    \end{minipage}
    \hfill
    \begin{minipage}[c]{.45\textwidth}
       \includegraphics[width=2.6in,height=2.03in]{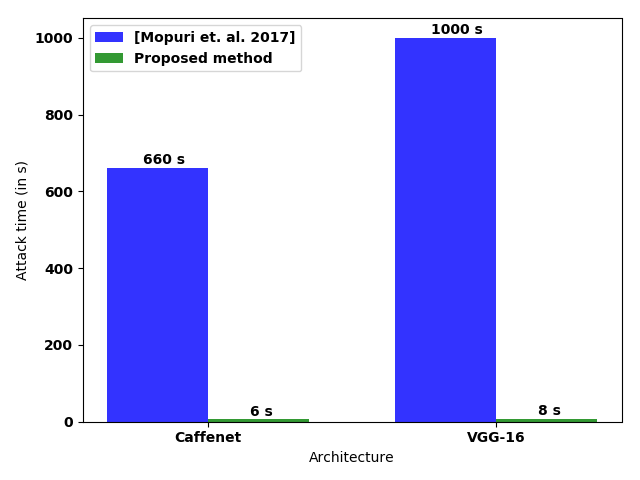}
     \begin{center}
    \caption{\footnotesize{Time needed to compute perturbation for different models over Imagenet.}}
    \label{boxplot1}
\end{center}
    \end{minipage}
\end{figure}   


\paragraph{Attack time:}
We compare the time needed to compute our adversarial perturbation with \cite{mopuri2017fast}. We use the code provided by the authors for this comparison. As shown in Fig. 2, our attack is orders of magnitude faster. We timed the algorithms on a GeForce GTX TITAN X GPU. We refer the reader to \citep{mopuri2017fast} for timing comparison between \citep{mopuri2017fast} and \citep{Moosavi-Dezfooli_2017_CVPR}. We don't compare with adversarial perturbation generator networks \citep{reddy2018nag,hayesuniversal} that can generate adversarial perturbation very fast but take much longer to be trained for a new target network.

\section{Discussion}
To gain understanding about the workings of our proposed perturbation, we perform various experiments and analyze its findings in this section.

\paragraph{Relationship between Proposed Universal Attack and Example Specific Adversarial Perturbations: } We find that the direction of proposed universal attack is aligned with the direction of example-specific adversarial examples (constructed via FGSM attack). We construct a matrix $P$ with each column as example-specific adversarial perturbations for MNIST test set. The singular values of $P$ decay quickly, as shown in Fig. 4A. This suggests that most example-specific perturbations reside in a small subspace spanned by first few singular vectors and any perturbation inside this subspace should be a good universal perturbation. In fact, a random vector within $L_{\infty}$ bound sampled from subspace spanned by the first 5 singular vectors obtains 36\% fooling rate on average. Note that computation of singular values and testing of attack success is done on disjoint splits of the test set. We found that perturbations computed using our method have high cosine similarity with the first few singular vectors, as shown in Fig. 4B (undefended model).

We also compute the matrix $P$ for an adversarially trained model on MNIST and observed several differences in properties of its singular values. The singular values decrease more slowly, as shown in Fig. 3A, and this suggests that the example-specific perturbations don't reside in a small subspace. A random vector within $L_{\infty}$ bound sampled from subspace spanned by the first 5 singular vectors only has 1.7\% fooling rate on average. These results show that directions of example-specific perturbations don't have shared direction for adversarially trained models and therefore a single vector can't have a high fooling rate. We would like to point out that adversarial training can be used to train a model robust to any attack and this weakness isn't specific to our attack.

\begin{figure}[t]
    \centering
    \begin{minipage}{0.45\textwidth}
        \centering
        \includegraphics[width=2.6in,height=2.3in]{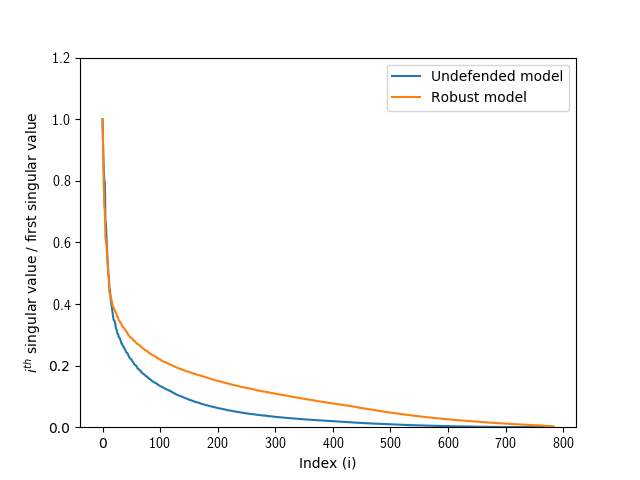}
\hspace{0.1cm}
\includegraphics[width=2.6in,height=2.3in]{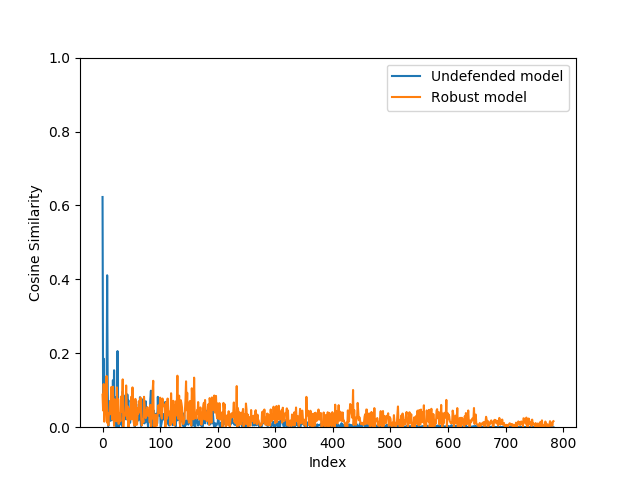}
        \caption{\footnotesize{A. (top) Ratio of $i^{th}$ singular value to first singular value of matrix $P$ containing example-wise adversarial perturbations. B. (bottom) Cosine similarity of our universal perturbation for class $0$ with singular vectors of matrix $P$.}}
    \end{minipage}\hfill
    \begin{minipage}{0.45\textwidth}
        \centering
        \includegraphics[width=2.6in,height=2.3in]{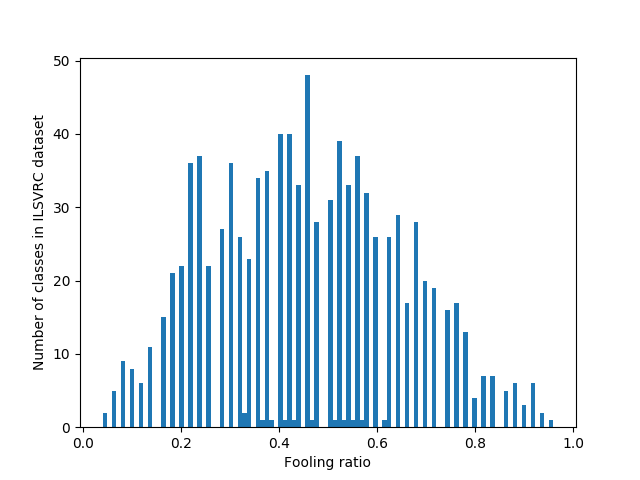}
\includegraphics[width=2.6in,height=2.3in]{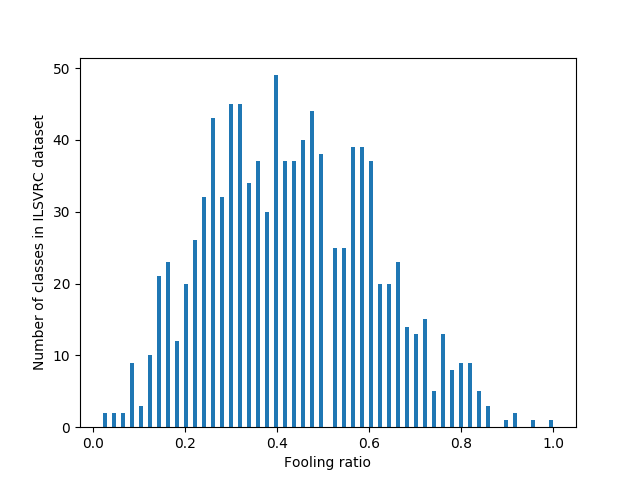}
        \caption{\footnotesize{Histograms showing number of classes with different fooling rates for VGG-16 and Resnet-18 (top to bottom). The average fooling rate for these models is $46.00\%$ and $43.57\%$ respectively.}}
    \end{minipage}
\end{figure}


\paragraph{Decision boundaries in deep neural networks are largely linear:} The decision boundary between class $i$ and $j$ for the classifier $f$ can be defined as:
\begin{equation}
    B_{ij} = \{\vec{z} : f_{i}(\vec{z}) = f_{j}(\vec{z})\}
\end{equation}
i.e. points which are equally likely to be classified in both $i_{th}$ and $j_{th}$ class.\\
We obtain samples of $B_{ij}$ for MNIST model using adversarial examples. For a data-point $x$ of class $i$ and its corresponding adversarial example belonging to class $j$ (obtained using targeted FGSM attack), we interpolate between the two points to get a data-point $\vec{z} \in B_{ij}$ which is equally likely to be classified in class $i$ or $j$. For the samples obtained in each pairwise decision boundary set $B_{ij}$ ($i,j\in {0,1,2,...,9}$ and $i\ne j$), we fit a hyper-plane and obtain its predicted $R^2$(Coefficient of determination). These $R^2$ values can be used to comment on the goodness of fit of the hyper-plane to the decision boundary, and thus a high $R^2$ means that the hyper-plane can fit the decision boundary with low error. We find that a significant number of pair-wise boundaries have high predicted $R^2$ (38 out of 81 class-wise boundaries have predicted $R^2 >= 0.9$). The results, thus suggest that the decision boundaries exhibit a linear nature and thus can be exploited by our attack.

\begin{figure*}[t]
\begin{center}
\includegraphics[width=1.25in,height=1.25in]{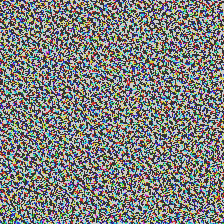}
\hspace{0.1cm}
\includegraphics[width=1.25in,height=1.25in]{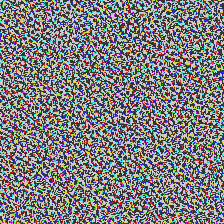}
\hspace{0.1cm}
\includegraphics[width=1.25in,height=1.25in]{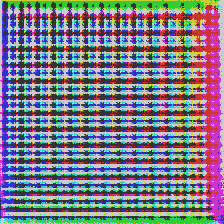}
\hspace{0.1cm}
\includegraphics[width=1.25in,height=1.25in]{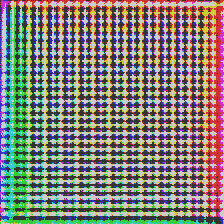}
\end{center}

\begin{center}
    \caption{(left to right): Adversarial perturbations for VGG-16 with image class
as horse-cart and table-lamp, adversarial perturbations for Resnet-18 with image class as horse-cart and table-lamp.}
\label{fig:exampleperturb2}
\end{center}
\end{figure*}

We also compute the $R^2$ values of class boundaries for an adversarially trained model and find that most of the $R^2$ values are close to $0$ (None of 81 class-wise boundaries have predicted $R^2 >= 0.9$). This shows that the linearity of decision boundaries breaks down for adversarially trained networks.

Previous adversarial attack methods \citep{goodfellow2014explaining} have also been motivated by linearity of deep neural networks.

\textbf{We find that some classes in ImageNet dataset are inherently harder to misclassify using universal adversarial perturbations.} In Table 1, we show fooling rates averaged over all classes for images in the ILSVRC validation set. We, now evaluate the fooling rate for each class separately to examine whether all classes are equally vulnerable to our attack or are some classes easier/harder to mis-classify. Fig. 5 is a frequency histogram of number of classes at different fooling rates. We observe high variance in fooling rates among different classes. For example, VGG-16 for ImageNet dataset has average fooling rate of 46.00\% but 77 classes have fooling rate below 20\% and 37 classes have fooling rate above 80\%. \\

We also check if the same classes are vulnerable to our attack across different architectures. We compare fooling rates for each class on VGG-16 and Resnet-18 architectures. The difference in fooling rate of each class on the two architectures, averaged over all classes is only 10.96\%. This shows that some classes in ILSVRC are inherently harder/easier to mis-classify than others.

\section{Conclusion}
We developed a new method of computing class-wise universal adversarial perturbations that does not require any training data. The method is particularly interesting due to its simplicity - universal adversarial perturbations can be directly computed as a linear function of weights of the neural network. We reported high attack rates across MNIST, CIFAR-10 and ImageNet datasets and found that the adversarial perturbations transfers well across models. We also perform some experiments that help in understanding how the proposed attack works.

Our results show that the proposed method leverages the linear nature of neural networks to construct class-wise universal perturbations. Our future work will focus on studying the attack method to uncover more insights into the vulnerabilities of deep neural networks to adversarial perturbations.



\bibliography{iclr2019_conference}

\begin{thebibliography}{37}
\providecommand{\natexlab}[1]{#1}
\providecommand{\url}[1]{\texttt{#1}}
\expandafter\ifx\csname urlstyle\endcsname\relax
  \providecommand{\doi}[1]{doi: #1}\else
  \providecommand{\doi}{doi: \begingroup \urlstyle{rm}\Url}\fi

\bibitem[Abadi et~al.(2016)Abadi, Barham, Chen, Chen, Davis, Dean, Devin,
  Ghemawat, Irving, Isard, et~al.]{abadi2016tensorflow}
Mart{\'\i}n Abadi, Paul Barham, Jianmin Chen, Zhifeng Chen, Andy Davis, Jeffrey
  Dean, Matthieu Devin, Sanjay Ghemawat, Geoffrey Irving, Michael Isard, et~al.
\newblock Tensorflow: a system for large-scale machine learning.
\newblock In \emph{OSDI}, 2016.

\bibitem[Akhtar \& Mian(2018)Akhtar and Mian]{reviewpaper2}
Naveed Akhtar and Ajmal Mian.
\newblock Threat of adversarial attacks on deep learning in computer vision: A
  survey.
\newblock \emph{IEEE Access}, 2018.

\bibitem[Athalye et~al.(2018)Athalye, Carlini, and
  Wagner]{athalye2018obfuscated}
Anish Athalye, Nicholas Carlini, and David Wagner.
\newblock Obfuscated gradients give a false sense of security: Circumventing
  defenses to adversarial examples.
\newblock \emph{arXiv preprint arXiv:1802.00420}, 2018.

\bibitem[Bhagoji et~al.(2017)Bhagoji, He, Li, and Song]{bhagoji2017exploring}
Arjun~Nitin Bhagoji, Warren He, Bo~Li, and Dawn Song.
\newblock Exploring the space of black-box attacks on deep neural networks.
\newblock \emph{arXiv preprint arXiv:1712.09491}, 2017.

\bibitem[Carlini \& Wagner(2017)Carlini and Wagner]{carlini2017towards}
Nicholas Carlini and David Wagner.
\newblock Towards evaluating the robustness of neural networks.
\newblock In \emph{IEEE Symposium on Security and Privacy}, 2017.

\bibitem[Carlini et~al.(2017)Carlini, Katz, Barrett, and
  Dill]{provable_certificate4}
Nicholas Carlini, Guy Katz, Clark Barrett, and David~L Dill.
\newblock Provably minimally-distorted adversarial examples.
\newblock \emph{arXiv preprint arXiv:1709.10207}, 2017.

\bibitem[Chatfield et~al.(2014)Chatfield, Simonyan, Vedaldi, and
  Zisserman]{chatfield2014return}
Ken Chatfield, Karen Simonyan, Andrea Vedaldi, and Andrew Zisserman.
\newblock Return of the devil in the details: Delving deep into convolutional
  nets.
\newblock \emph{arXiv preprint arXiv:1405.3531}, 2014.

\bibitem[Engstrom et~al.(2018)Engstrom, Ilyas, and Athalye]{alp2018broken}
Logan Engstrom, Andrew Ilyas, and Anish Athalye.
\newblock Evaluating and understanding the robustness of adversarial logit
  pairing.
\newblock \emph{NeurIPS SECML}, 2018.

\bibitem[Goodfellow et~al.(2015)Goodfellow, Shlens, and
  Szegedy]{goodfellow2014explaining}
Ian~J Goodfellow, Jonathon Shlens, and Christian Szegedy.
\newblock Explaining and harnessing adversarial examples.
\newblock In \emph{ICLR}, 2015.

\bibitem[Hayes \& Danezis(2017)Hayes and Danezis]{hayesuniversal}
Jamie Hayes and George Danezis.
\newblock Learning universal adversarial perturbations with generative models.
\newblock \emph{arXiv preprint arXiv:1708.05207}, 2017.

\bibitem[He et~al.(2016)He, Zhang, Ren, and Sun]{he2016deep}
Kaiming He, Xiangyu Zhang, Shaoqing Ren, and Jian Sun.
\newblock Deep residual learning for image recognition.
\newblock In \emph{CVPR}, 2016.

\bibitem[Ilyas et~al.(2018{\natexlab{a}})Ilyas, Engstrom, Athalye, and
  Lin]{ilyas2018black}
Andrew Ilyas, Logan Engstrom, Anish Athalye, and Jessy Lin.
\newblock Black-box adversarial attacks with limited queries and information.
\newblock \emph{arXiv preprint arXiv:1804.08598}, 2018{\natexlab{a}}.

\bibitem[Ilyas et~al.(2018{\natexlab{b}})Ilyas, Engstrom, and
  Madry]{ilyas2018prior}
Andrew Ilyas, Logan Engstrom, and Aleksander Madry.
\newblock Prior convictions: Black-box adversarial attacks with bandits and
  priors.
\newblock \emph{arXiv preprint arXiv:1807.07978}, 2018{\natexlab{b}}.

\bibitem[Jia et~al.(2014)Jia, Shelhamer, Donahue, Karayev, Long, Girshick,
  Guadarrama, and Darrell]{jia2014caffe}
Yangqing Jia, Evan Shelhamer, Jeff Donahue, Sergey Karayev, Jonathan Long, Ross
  Girshick, Sergio Guadarrama, and Trevor Darrell.
\newblock Caffe: Convolutional architecture for fast feature embedding.
\newblock In \emph{
  Multimedia ACM}. ACM, 2014.

\bibitem[Khrulkov \& Oseledets(2018)Khrulkov and
  Oseledets]{art_singular_2018_CVPR}
Valentin Khrulkov and Ivan Oseledets.
\newblock Art of singular vectors and universal adversarial perturbations.
\newblock In \emph{
  Recognition ( CVPR}, July 2018.

\bibitem[Kolter \& Wong(2017)Kolter and Wong]{kolter2017provable}
J~Zico Kolter and Eric Wong.
\newblock Provable defenses against adversarial examples via the convex outer
  adversarial polytope.
\newblock \emph{arXiv preprint arXiv:1711.00851}, 2017.

\bibitem[Krizhevsky(2009)]{cifar_dataset}
Alex Krizhevsky.
\newblock Learning multiple layers of features from tiny images.
\newblock 2009.

\bibitem[Kurakin et~al.(2017)Kurakin, Goodfellow, and
  Bengio]{kurakin2016adversarial}
Alexey Kurakin, Ian Goodfellow, and Samy Bengio.
\newblock Adversarial examples in the physical world.
\newblock \emph{ICLR Workshop}, 2017.

\bibitem[Lecun et~al.(1989)Lecun, Boser, Denker, Henderson, Howard, Hubbard,
  and Jackel]{mnist_dataset}
Yan Lecun, B.~Boser, J.S. Denker, D.~Henderson, R.E. Howard, W.~Hubbard, and
  L.D. Jackel.
\newblock Backpropagation applied to handwritten zip code recognition, 1989.

\bibitem[Liu et~al.(2016)Liu, Chen, Liu, and Song]{liu2016delving}
Yanpei Liu, Xinyun Chen, Chang Liu, and Dawn Song.
\newblock Delving into transferable adversarial examples and black-box attacks.
\newblock \emph{arXiv preprint arXiv:1611.02770}, 2016.

\bibitem[Madry et~al.(2018)Madry, Makelov, Schmidt, Tsipras, and
  Vladu]{madry2017towards}
Aleksander Madry, Aleksandar Makelov, Ludwig Schmidt, Dimitris Tsipras, and
  Adrian Vladu.
\newblock Towards deep learning models resistant to adversarial attacks.
\newblock In \emph{ICLR}, 2018.

\bibitem[Moosavi-Dezfooli et~al.(2016)Moosavi-Dezfooli, Fawzi, and
  Frossard]{moosavi2016deepfool}
Seyed-Mohsen Moosavi-Dezfooli, Alhussein Fawzi, and Pascal Frossard.
\newblock Deepfool: a simple and accurate method to fool deep neural networks.
\newblock In \emph{
  Pattern Recognition CVPR}, 2016.

\bibitem[Moosavi-Dezfooli et~al.(2017{\natexlab{a}})Moosavi-Dezfooli, Fawzi,
  Fawzi, and Frossard]{Moosavi-Dezfooli_2017_CVPR}
Seyed-Mohsen Moosavi-Dezfooli, Alhussein Fawzi, Omar Fawzi, and Pascal
  Frossard.
\newblock Universal adversarial perturbations.
\newblock In \emph{
  Recognition CVPR}, July 2017{\natexlab{a}}.

\bibitem[Moosavi-Dezfooli et~al.(2017{\natexlab{b}})Moosavi-Dezfooli, Fawzi,
  Fawzi, Frossard, and Soatto]{moosavi2017analysis}
Seyed-Mohsen Moosavi-Dezfooli, Alhussein Fawzi, Omar Fawzi, Pascal Frossard,
  and Stefano Soatto.
\newblock Analysis of universal adversarial perturbations.
\newblock \emph{CoRR}, abs/1705.09554, 2017{\natexlab{b}}.

\bibitem[Mopuri et~al.(2017)Mopuri, Garg, and Babu]{mopuri2017fast}
Konda~Reddy Mopuri, Utsav Garg, and R~Venkatesh Babu.
\newblock Fast feature fool: A data independent approach to universal
  adversarial perturbations.
\newblock In \emph{BMVC}, 2017.

\bibitem[Papernot et~al.(2017)Papernot, McDaniel, Goodfellow, Jha, Celik, and
  Swami]{papernot2017practical}
Nicolas Papernot, Patrick McDaniel, Ian Goodfellow, Somesh Jha, Z~Berkay Celik,
  and Ananthram Swami.
\newblock Practical black-box attacks against machine learning.
\newblock In \emph{
  and Communications Security ACM}, 2017.

\bibitem[Paszke et~al.(2017)Paszke, Gross, Chintala, Chanan, Yang, DeVito, Lin,
  Desmaison, Antiga, and Lerer]{paszke2017automatic}
Adam Paszke, Sam Gross, Soumith Chintala, Gregory Chanan, Edward Yang, Zachary
  DeVito, Zeming Lin, Alban Desmaison, Luca Antiga, and Adam Lerer.
\newblock Automatic differentiation in pytorch.
\newblock 2017.

\bibitem[Poursaeed et~al.(2018)Poursaeed, Katsman, Gao, and
  Belongie]{poursaeed2018generative}
Omid Poursaeed, Isay Katsman, Bicheng Gao, and Serge Belongie.
\newblock Generative adversarial perturbations.
\newblock In \emph{Proceedings of the IEEE Conference on Computer Vision and
  Pattern Recognition}, pp.\  4422--4431, 2018.

\bibitem[Raghunathan et~al.(2018)Raghunathan, Steinhardt, and
  Liang]{raghunathan2018certified}
Aditi Raghunathan, Jacob Steinhardt, and Percy Liang.
\newblock Certified defenses against adversarial examples.
\newblock \emph{arXiv preprint arXiv:1801.09344}, 2018.

\bibitem[Reddy~Mopuri et~al.(2018)Reddy~Mopuri, Ojha, Garg, and
  Venkatesh~Babu]{reddy2018nag}
Konda Reddy~Mopuri, Utkarsh Ojha, Utsav Garg, and R~Venkatesh~Babu.
\newblock Nag: Network for adversary generation.
\newblock In \emph{CVPR}, 2018.

\bibitem[Russakovsky et~al.(2015)Russakovsky, Deng, Su, Krause, Satheesh, Ma,
  Huang, Karpathy, Khosla, Bernstein, et~al.]{russakovsky2015imagenet}
Olga Russakovsky, Jia Deng, Hao Su, Jonathan Krause, Sanjeev Satheesh, Sean Ma,
  Zhiheng Huang, Andrej Karpathy, Aditya Khosla, Michael Bernstein, et~al.
\newblock Imagenet large scale visual recognition challenge.
\newblock \emph{IJCV}, 2015.

\bibitem[Simonyan \& Zisserman(2015)Simonyan and Zisserman]{simonyan2014very}
Karen Simonyan and Andrew Zisserman.
\newblock Very deep convolutional networks for large-scale image recognition.
\newblock In \emph{ICLR}, 2015.

\bibitem[Szegedy et~al.(2014)Szegedy, Zaremba, Sutskever, Bruna, Erhan,
  Goodfellow, and Fergus]{szegedy2013intriguing}
Christian Szegedy, Wojciech Zaremba, Ilya Sutskever, Joan Bruna, Dumitru Erhan,
  Ian Goodfellow, and Rob Fergus.
\newblock Intriguing properties of neural networks.
\newblock \emph{ICLR}, 2014.

\bibitem[Tsipras et~al.(2018)Tsipras, Santurkar, Engstrom, Turner, and
  Madry]{tsipras2018there}
Dimitris Tsipras, Shibani Santurkar, Logan Engstrom, Alexander Turner, and
  Aleksander Madry.
\newblock There is no free lunch in adversarial robustness (but there are
  unexpected benefits).
\newblock \emph{arXiv preprint arXiv:1805.12152}, 2018.

\bibitem[Uesato et~al.(2018)Uesato, O'Donoghue, Oord, and
  Kohli]{uesato2018adversarial}
Jonathan Uesato, Brendan O'Donoghue, Aaron van~den Oord, and Pushmeet Kohli.
\newblock Adversarial risk and the dangers of evaluating against weak attacks.
\newblock \emph{arXiv preprint arXiv:1802.05666}, 2018.

\bibitem[Wong et~al.(2018)Wong, Schmidt, Metzen, and Kolter]{wong2018scaling}
Eric Wong, Frank Schmidt, Jan~Hendrik Metzen, and J~Zico Kolter.
\newblock Scaling provable adversarial defenses.
\newblock In \emph{NIPS}, 2018.

\bibitem[Yuan et~al.(2019)Yuan, He, Zhu, and Li]{reviewpaper}
Xiaoyong Yuan, Pan He, Qile Zhu, and Xiaolin Li.
\newblock Adversarial examples: Attacks and defenses for deep learning.
\newblock \emph{IEEE transactions on neural networks and learning systems},
  2019.

\end{thebibliography}
\bibliographystyle{iclr2019_conference}

\end{document}